\documentclass[11pt,a4paper]{article}
\usepackage[hyperref]{acl2019}
\usepackage{times,latexsym}
\usepackage[shortcuts]{extdash}
\usepackage{url}
\usepackage[T1]{fontenc}
\usepackage{amssymb}
\usepackage{amsmath}
\usepackage{bbm}
\usepackage{breqn}
\usepackage{algorithm}
\usepackage{algpseudocode}
\usepackage{mathptmx}
\usepackage{multicol}
\usepackage{tikz}
\usepackage{booktabs}
\usepackage{tabularx}
\usepackage{dcolumn}
\newcolumntype{d}[1]{D{.}{.}{#1}}
\usetikzlibrary{shapes.geometric}
\usetikzlibrary{arrows.meta}

\DeclareMathAlphabet{\mathcal}{OMS}{cmsy}{m}{n}
\DeclareSymbolFont{matha}{OML}{txmi}{m}{it}
\DeclareMathSymbol{\varv}{\mathord}{matha}{118}
\newcolumntype{R}{>{\raggedleft\arraybackslash}X}

\usepackage{url}

\usepackage{xspace,mfirstuc,tabulary}

\aclfinalcopy 
\def\aclpaperid{912} 

\title{Understanding Undesirable Word Embedding Associations}

\author{Kawin Ethayarajh, David Duvenaud$^\dagger$, Graeme Hirst \\
  University of Toronto\\
  $^\dagger$Vector Institute \\
  {\tt \{kawin, duvenaud, gh\}@cs.toronto.edu} \\
}

\date{}

\begin{document}
\maketitle
\begin{abstract}
  Word embeddings are often criticized for capturing undesirable word associations such as gender stereotypes. However, methods for measuring and removing such biases remain poorly understood. We show that for any embedding model that implicitly does matrix factorization, debiasing vectors \emph{post hoc} using subspace projection \citep{bolukbasi2016man} is, under certain conditions, equivalent to training on an unbiased corpus. We also prove that WEAT, the most common association test for word embeddings, systematically overestimates bias. Given that the subspace projection method is provably effective, we use it to derive a new measure of association called the \emph{relational inner product association} (RIPA). Experiments with RIPA reveal that, on average, skipgram with negative sampling (SGNS) does not make most words any more gendered than they are in the training corpus. However, for gender-stereotyped words, SGNS actually amplifies the gender association in the corpus.
\end{abstract}

\section{Introduction}

A common criticism of word embeddings is that they capture undesirable associations in vector space. In addition to gender-appropriate analogies such as \emph{king:queen::man:woman}, stereotypical analogies such as \emph{doctor:nurse::man:woman} also hold in SGNS embedding spaces \citep{bolukbasi2016man}. \citet{caliskan2017semantics} created an association test for word vectors called WEAT, which uses cosine similarity to measure how associated words are with respect to two sets of attribute words (e.g., `male' vs.\ `female'). For example, they claimed that science-related words were significantly more associated with male attributes and art-related words with female ones. Since these associations are socially undesirable, they were described as gender bias.

Despite these remarkable findings, such undesirable word associations remain poorly understood. For one, what causes them -- is it biased training data, the embedding model itself, or just noise? Why should WEAT be the test of choice for measuring associations in word embeddings? \citet{bolukbasi2016man} found that word vectors could be debiased by defining a ``bias subspace'' in the embedding space and then subtracting from each vector its projection on this subspace. But what theoretical guarantee is there that this method actually debiases vectors? 

In this paper, we answer several of these open questions. We begin by proving that for any embedding model that implicitly does matrix factorization (e.g., GloVe, SGNS), debiasing vectors \emph{post hoc} via subspace projection is, under certain conditions, equivalent to training on an unbiased corpus without reconstruction error. We find that contrary to what \citet{bolukbasi2016man} suggested, word embeddings \emph{should not} be normalized before debiasing, as vector length can contain important information \cite{ethayarajh2018towards}. To guarantee unbiasedness, the bias subspace should also be the span -- rather than a principal component -- of the vectors used to define it. If applied this way, the subspace projection method can be used to provably debias SGNS and GloVe embeddings with respect to the word pairs that define the bias subspace.

Using this notion of a ``bias subspace'', we then prove that WEAT, the most common association test for word embeddings, has theoretical flaws that cause it to systematically overestimate bias. At least for SGNS and GloVe, it implicitly requires the two sets of attribute words (e.g., `male' vs.\ `female') to occur with equal frequency in the training corpus; when they do not, even gender-neutral words can be classified as gender-biased, for example. The outcome of a WEAT test can also be easily manipulated by contriving the attribute word sets, allowing virtually any word -- even a gender-neutral one such as `door' -- to be classified as male- or female-biased relative to another gender-neutral word.

Given that subspace projection removal provably debiases embeddings, we use it to derive a new measure of association in word embeddings called the \emph{relational inner product association} (RIPA). Given a set of ordered word pairs (e.g., \{(`man', `woman'), (`male', `female')\}), we take the first principal component of all the difference vectors, which we call the \emph{relation vector} $\vec{b}$. In \citeauthor{bolukbasi2016man}'s terminology, $\vec{b}$ would be a one-dimensional bias subspace. Then, for a word vector $\vec{w}$, the relational inner product is simply $\langle \vec{w}, \vec{b} \rangle$. Because RIPA is intended for embedding models that implicitly do matrix factorization, it has an information theoretic interpretation. This allows us to directly compare the actual word association in embedding space with what we would expect the word association to be, given the training corpus. Making such comparisons yields several novel insights:
\begin{enumerate}
    \item SGNS does not, \emph{on average}, make the vast majority of words any more gendered in the vector space than they are in the training corpus; individual words may be slightly more or less gendered due to reconstruction error. However, for words that are gender-stereotyped (e.g., `nurse') or gender-specific by definition (e.g., `queen'), SGNS amplifies the gender association in the training corpus. 
    \item To use the subspace projection method, one must have prior knowledge of which words are gender-specific by definition, so that they are not also debiased. Debiasing all vectors can preclude gender-appropriate analogies such as \emph{king:queen::man:woman} from holding in the embedding space. In contrast to the supervised method proposed by \citet{bolukbasi2016man} for identifying these gender-specific words, we introduce an unsupervised method. Ours is much more effective at preserving gender-appropriate analogies and precluding gender-biased ones.
\end{enumerate}
To allow a fair comparison with prior work, our experiments in this paper focus on gender association. However, our claims extend to other types of word associations as well, which we leave as future work.

\section{Related Work}

\paragraph{Word Embeddings} Word embedding models generate distributed representations of words in a low-dimensional continuous space. This is generally done using: (a) neural networks that learn embeddings by predicting the contexts words appear in, or vice-versa \citep{bengio2003neural,mikolov2013distributed,collobert2008unified}; (b) low-rank approximations of word-context matrices containing a co-occurrence statistic \citep{landauer1997solution,levy2014neural}. The objective of SGNS is to maximize the probability of observed word-context pairs and to minimize the probability of $k$ randomly sampled negative examples. Though no co-occurrence statistics are explicitly calculated, \citet{levy2014neural} proved that SGNS is implicitly factorizing a word-context PMI matrix shifted by $- \log k$. Similarly, GloVe implicitly factorizes a log co-occurrence count matrix \cite{pennington2014glove}.

\paragraph{Word Analogies} A word analogy \emph{a:b::x:y} asserts that ``\emph{a is to b as x is to y}'' and holds in the embedding space iff $\vec{a} + ( \vec{y} - \vec{x}) = \vec{b}$. \citet{ethayarajh2018towards} proved that for GloVe and SGNS, \emph{a:b::x:y} holds exactly in an embedding space with no reconstruction error iff the words are coplanar and the co-occurrence shifted PMI is the same for each word pair and across both word pairs. Word analogies are often used to signify that semantic and syntactic properties of words (e.g., verb tense, gender) can be captured as linear relations.

\paragraph{Measuring Associations} \citet{caliskan2017semantics} proposed what is now the most commonly used association test for word embeddings. The word embedding association test (WEAT) uses cosine similarity to measure how associated two given sets of target words are with respect to two sets of attribute words (e.g., `male' vs.\ `female'). For example, \citet{caliskan2017semantics} claimed that science-related words are more associated with `male' than `female' attributes compared to art-related words, and that this was statistically significant. However, aside from some intuitive results (e.g., that female names are associated with female attributes), there is little evidence that WEAT is a good measure of association. 

\paragraph{Debiasing Embeddings} \citet{bolukbasi2016man} claimed that the existence of stereotypical analogies such as \emph{doctor:nurse::man:woman} constituted gender bias. To prevent such analogies from holding in the vector space, they subtracted from each biased word vector its projection on a ``gender bias subspace''. This subspace was defined by the first $m$ principal components for ten gender relation vectors (e.g., $\vec{\textit{man}} - \vec{\textit{woman}}$). Each debiased word vector was thus orthogonal to the gender bias subspace and its projection on the subspace was zero. While this subspace projection method precluded gender-biased analogies from holding in the embedding space, \citet{bolukbasi2016man} did not provide any theoretical guarantee that the vectors were unbiased (i.e., equivalent to vectors that would be obtained from training on a gender-agnostic corpus with no reconstruction error). Other work has tried to learn gender-neutral embeddings from scratch \citep{zhao2018learning}, despite this approach requiring custom changes to the objective of each embedding model.

\section{Provably Debiasing Embeddings}

Experiments by \citet{bolukbasi2016man} found that debiasing word embeddings using the subspace projection method precludes gender-biased analogies from holding. However, as we noted earlier, despite this method being intuitive, there is no theoretical guarantee that the debiased vectors are perfectly unbiased or that the debiasing method works for embedding models other than SGNS. In this section, we prove that for any embedding model that does implicit matrix factorization (e.g., GloVe, SGNS), debiasing embeddings \emph{post hoc} using the subspace projection method is, under certain conditions, equivalent to training on a perfectly unbiased corpus without reconstruction error. 

\paragraph{Definition 1} \emph{Let $M$ denote the symmetric word-context matrix for a given training corpus that is implicitly or explicitly factorized by the embedding model. Let $S$ denote a set of word pairs. A word $w$ is \emph{unbiased} with respect to $S$ iff $\forall\, (x,y) \in S, M_{w,x} = M_{w,y}$. $M$ is \emph{unbiased} with respect to $S$ iff $\forall\, w \not \in S$, $w$ is unbiased. A word $w$ or matrix $M$ is \emph{biased} wrt $S$ iff it is not unbiased wrt $S$.}

Note that Definition 1 does not make any distinction between socially acceptable and socially unacceptable associations. A word that is gender-specific by definition and a word that is gender-biased due to stereotypes would both be considered biased by Definition 1, although only the latter is undesirable. For example, by Definition 1, `door' would be unbiased with respect to the set \{(`male', `female')\} iff the entries for $M_{\text{door}, \text{male}}$ and $M_{\text{door}, \text{female}}$ were interchangeable. The entire corpus would be unbiased with respect to the set iff $M_{\textit{w}, \text{male}}$ and $M_{\textit{w}, \text{female}}$ were interchangeable for any word $w$. Since $M$ is a word-context matrix containing a co-occurrence statistic, unbiasedness effectively means that the elements for $(w, \text{`male'})$ and $(w, \text{`female'})$ in $M$ can be switched without any impact on the embeddings. $M$ is factorized into a word matrix $W$ and context matrix $C$ such that $WC^T = M$, with the former giving us our word embeddings.

\paragraph{Debiasing Theorem} \emph{For a set of word pairs $S$, let the bias subspace $B = \text{span}(\{ \vec{x} - \vec{y}\, |\, (x,y) \in S\})$. For every word $w \not \in S$, let $\vec{w_d} \triangleq \vec{w} - \text{proj}_B \vec{w}$. The  reconstructed word-context matrix $W_d C^T = M_d$ is unbiased with respect to $S$.}

\paragraph{Proof of Theorem} When there is no reconstruction error, we know from Definition 1 that a word $w$ is unbiased with respect to a set of word pairs $S$ iff $\forall\, (x,y) \in S$
\begin{equation}
\begin{split}
M_{w,x} = M_{w,y} &\iff \left< \vec{w}, \vec{x_c} \right> = \left< \vec{w}, \vec{y_c} \right> \\ &\iff \left< \vec{w}, \vec{x_c} - \vec{y_c} \right> = 0
\end{split}
\end{equation}

From Lemma 2 of \citet{ethayarajh2018towards}, we also know that under perfect reconstruction, and assuming that the mapping from word to context space preserves relative geometry globally and contains no reflections, $\exists\ \lambda \in \mathbb{R}, C = \lambda W$. For a detailed explanation, we refer the reader to the proof of that lemma. All word vectors must therefore lie in the same eigenspace, with eigenvalue $\lambda$. This implies that for any word $w$ and any $(x,y) \in S$, 
\begin{equation}
    \exists\ \lambda \in \mathbb{R}, \left< \vec{w}, \vec{x_c} - \vec{y_c} \right> = \lambda \left< \vec{w}, \vec{x} - \vec{y} \right>
    \label{lemma2}
\end{equation}
Each debiased word vector $w_d$ is orthogonal to the bias subspace in the word embedding space, so $\forall\, (x,y) \in S, \left< \vec{w_d}, \vec{x} - \vec{y} \right> = 0$. In conjunction with (\ref{lemma2}), this implies that $\forall\, (x,y) \in S, \lambda \left< \vec{w_d}, \vec{x} - \vec{y} \right> = \left< \vec{w_d}, \vec{x_c} - \vec{y_c} \right> = 0$. This means that if a debiased word $w$ is represented with vector $\vec{w}_d$ instead of $\vec{w}$, it is unbiased with respect to $S$ by Definition 1. This implies that the co-occurrence matrix $M_d$ that is reconstructed using the debiased word matrix $W_d$ is also unbiased with respect to $S$.

The subspace projection method is therefore far more powerful than initially stated in \citet{bolukbasi2016man}: not only can it be applied to any embedding model that implicitly does matrix factorization (e.g., GloVe, SGNS), but debiasing word vectors in this way is equivalent to training on a perfectly unbiased corpus when there is no reconstruction error. However, word vectors \emph{should not} be normalized prior to debiasing, since the matrix that is factorized by the embedding model cannot necessarily be reconstructed with normalized embeddings. Unbiasedness with respect to word pairs $S$ is also only guaranteed when the bias subspace $B = \text{span}(\{ \vec{x} - \vec{y} | (x, y) \in S \})$.

Because we define unbiasedness with respect to a set of word pairs, we cannot make any claims about word pairs outside that set. For example, consider the set $S = \{(\text{`man'}, \text{`woman'})\}$. If we define a bias subspace using $S$ and use it to debias $\vec{w}$, we can only say definitively that $\vec{w}$ is unbiased with respect to $S$. We cannot claim, for example, that $\vec{w}$ is also unbiased with respect to $\{(\text{`policeman'}, \text{`policewoman'})\}$, because it is possible that $\vec{\textit{policewoman}} - \vec{\textit{policeman}} \not= \vec{\textit{woman}} - \vec{\textit{man}}$. Debiasing $\vec{w}$ with respect to a non-exhaustive set of gender-defining word pairs is not equivalent to erasing all vestiges of gender from $\vec{w}$. This may explain why it is still possible to cluster words by gender after debiasing them using a handful of gender-defining word pairs \citep{gonen2019lipstick}.

\section{The Flaws of WEAT}

Given \emph{attribute word sets} $X$ and $Y$ (e.g., \{`male', `man'\} vs.\ \{`female', `woman'\}), WEAT uses a cosine similarity-based measurement to capture whether two \emph{target word sets} have the same relative association to both sets of attribute words. At the heart of WEAT is the statistic $s(w, X, Y)$, which "measures the association of [a word] $w$ with the attribute" \citep{caliskan2017semantics}:
\begin{equation}
    s(w, X, Y) = \mathbb{E}_{X} \cos(\vec{w}, \vec{x}) - \mathbb{E}_{Y} \cos(\vec{w}, \vec{y})
\end{equation}
The normalized difference between the mean values of $s(w, X, Y)$ across the two target word sets is called the \emph{effect size}. For the sake of simplicity, we consider the case where both attribute word sets contain a single word (i.e., $X = \{x\}, Y = \{y\}$).

\paragraph{Proposition 1} \emph{Let $X = \{x\}, Y = \{y\}$, and $w$ be unbiased with respect to $\{(x,y)\}$ by Definition 1. According to WEAT, an SGNS vector $\vec{w}$ is equally associated with $X$ and $Y$ under perfect reconstruction iff $p(x) = p(y)$.}

Both theoretical and empirical work have found the squared word embedding norm to be linear in the log probability of the word. \citep{arora2016latent,ethayarajh2018towards}. Where $\alpha_1, \alpha_2 \in \mathbb{R}$, $w$ is then equally associated with $X$ and $Y$ if
\begin{equation}
\begin{split}
    0 &= \cos(\vec{w}, \vec{x}) - \cos(\vec{w}, \vec{y}) \\
    &= \frac{1}{ \|\vec{w}\|_2} \left( \frac{\left< \vec{w}, \vec{x} \right>}{\| \vec{x} \|_2} - \frac{ \left< \vec{w}, \vec{y} \right>}{\| \vec{y} \|_2} \right) \\
    &= \frac{\left< \vec{w}, \vec{x} \right>}{\sqrt{ \alpha_1 \log p(x) + \alpha_2 } } - \frac{\left< \vec{w}, \vec{y} \right>}{ \sqrt{ \alpha_1 \log p(y) + \alpha_2 }}
\end{split}
\label{prop1}
\end{equation}
Under the conditions used in the proof of the Debiasing Theorem, $w$ is unbiased with respect to the set $\{(x,y)\}$ iff $\left< \vec{w}, \vec{x} \right> = \left< \vec{w}, \vec{y} \right>$. Therefore (\ref{prop1}) holds iff $p(x) = p(y)$. Thus for $w$ to be equally associated with both sets of attribute words, not only must $w$ be unbiased with respect to $\{(x,y)\}$ by Definition 1, but words $x$ and $y$ must also occur with equal frequency in the corpus. Despite this being implicitly required, it was not stated as a requirement in \citet{caliskan2017semantics} for using WEAT. If the embedding model were GloVe instead of SGNS, this requirement would still apply, since GloVe implicitly factorizes a log co-occurrence count matrix \cite{pennington2014glove} while SGNS implicitly factorizes the shifted PMI matrix \cite{levy2014neural}. 

This, in turn, means that the test statistic and effect size of WEAT can be non-zero even when each set of target words is unbiased with respect to the attribute words. In practice, this issue often goes unnoticed because each word in the attribute set, at least for gender association, has a counterpart that appears with roughly equal frequency in most training corpora (e.g., `man' vs.\ `woman', `boy' vs.\ `girl'). However, this is not guaranteed to hold, especially for more nebulous attribute sets (e.g., `pleasant' vs.\ `unpleasant' words).

\begin{table*}[t]
    \centering 
    \small
    \begin{tabularx}{\textwidth}{Xlcccl}
        \toprule Target Word Sets & Attribute Word Sets & Test Statistic & Effect Size & $p$-value & Outcome (WEAT)\\
        \midrule
 &  \{masculine\}  vs. \{feminine\} & \ \ \ 0.021 & \ \ \ 2.0 & 0.0 & more male-associated \\
\{door\} vs. \{curtain\} &  \{girlish\} vs. \{boyish\}  & $-$0.042 & $-$2.0 & 0.5 & inconclusive \\
& \{woman\}  vs. \{man\}  & \ \ \ 0.071 & \ \ \ 2.0 & 0.0 & more female-associated \\ \midrule

 &  \{masculine\}  vs. \{feminine\}  & \ \ \ 0.063 & \ \ \ 2.0 & 0.0 & more male-associated \\
 \{dog\} vs. \{cat\} &  \{actress\} vs. \{actor\}  & $-$0.075 & $-$2.0 & 0.5 & inconclusive \\
 & \{womanly\} vs. \{manly\}  & \ \ \ 0.001 & \ \ \ 2.0 & 0.0 & more female-associated \\ \midrule
 
 &  \{masculine\}  vs. \{feminine\}  & \ \ \ 0.017 & \ \ \ 2.0 & 0.0 & more male-associated \\
 \{bowtie\} vs. \{corsage\} & \{woman\} vs. \{masculine\}  & $-$0.071 & $-$2.0 & 0.5 & inconclusive \\
 & \{girly\} vs. \{masculine\}  & \ \ \ 0.054 & \ \ \ 2.0 & 0.0 & more female-associated \\  \bottomrule
        
    \end{tabularx}
    \caption{By contriving the male and female attribute words, we can easily manipulate WEAT to claim that a given target word is more female-biased or male-biased than another. For example, in the top row, $\vec{\textit{door}}$ is more male-associated than $\vec{\textit{curtain}}$ when the attribute words are `masculine' and `feminine', but it is more female-associated when the attribute words are `woman' and `man'. In both cases, the associations are highly statistically significant.}
    \label{tab:contrived_attributes}
\end{table*}

\paragraph{Proposition 2} \emph{Let $X = \{x\}, Y = \{y\}$, and the target word sets be $T_1 = \{w_1\}, T_2 = \{w_2\}$. Regardless of what the target words are, the effect size of their association with $X$ and $Y$ is maximal in one direction, according to WEAT.}

In this scenario, the effect size of the association is 2 (i.e., the maximum) in one of the two directions: either $w_1$ is more associated with $X$ than $Y$, or $w_2$ is. This is because the numerator of the effect size is the difference between $s(w_1, X, Y)$ and $s(w_2, X, Y)$, while the denominator is the standard deviation of $\{ s(w, X, Y) | w \in T_1 \cup T_2 \}$, which simplifies to $\sqrt{(s(w_1, X, Y) - s(w_2, X, Y))^2/4}$. This means that the effect size is necessarily 2 in one direction and $-$2 in the other; it is at its maximum regardless of how small individual similarities are. 

This also means that we can contrive the attribute word sets to achieve a desired outcome. For example, when the attribute word sets are \{`masculine'\} and \{`feminine'\}, $\vec{\textit{door}}$ is significantly more male-associated than $\vec{\textit{curtain}}$. When the attribute sets are  \{`woman'\} and \{`man'\}, the opposite is true: $\vec{\textit{door}}$ is significantly more female-associated than $\vec{\textit{curtain}}$. In Table \ref{tab:contrived_attributes}, we provide more examples of how we can easily contrive the attribute sets to claim, with high statistical significance, that a given target word is more female-biased or male-biased than another. Conversely, we can also manipulate the attribute sets to claim that an association is not statistically significant ($p = 0.5$), despite a large effect size. 

Broadly speaking, cosine similarity is a useful measure of vector similarity and hypothesis tests are useful for testing sample differences. Because of this, WEAT seems to be an intuitive measure. However, as shown in Propositions 1 and 2, there are two key theoretical flaws to WEAT that cause it to overestimate the degree of association and ultimately make it an inappropriate metric for word embeddings. The only other metric of note quantifies association as $|\cos(\vec{w}, \vec{b})|^c$, where $\vec{b}$ is the bias subspace and $c \in \mathbb{R}$ the ``strictness'' of the measurement \cite{bolukbasi2016man}. For the same reason discussed in Proposition 1, this measure can also overestimate the degree of association. 

\section{Relational Inner Product Association}

Given the theoretical flaws of WEAT, we derive a new measure of word embedding association using the subspace projection method, which can provably debias embeddings (section 3). 

\paragraph{Definition 2} \emph{The \emph{relational inner product association} $\beta(\vec{w}; \vec{b})$ of a word vector $\vec{w} \in V$ with respect to a relation vector $\vec{b} \in V$ is $\langle \vec{w}, \vec{b} \rangle$. Where $S$ is a non-empty set of ordered word pairs $(x,y)$ that define the association, $\vec{b}$ is the first principal component of $\{ \vec{x} - \vec{y}\ |\ (x,y) \in S \}$.}

Our metric, the \emph{relational inner product association} (RIPA), is simply the inner product of a relation vector describing the association and a given word vector in the same embedding space. To use the terminology in \citet{bolukbasi2016man}, RIPA is the scalar projection of a word vector onto a one-dimensional bias subspace defined by the unit vector $\vec{b}$. In their experiments, \citet{bolukbasi2016man} defined $\vec{b}$ as the first principal component for a set of gender difference vectors (e.g., $\vec{\textit{man}} - \vec{\textit{woman}}$). This would be the means of deriving $\vec{b}$ for RIPA as well. 

For the sake of interpretability, we do not define $\vec{b}$ as the span of difference vectors, as would be required if one were using $\vec{b}$ to provably debias words with respect to $S$ (see section 3). When $\vec{b}$ is a vector, the sign of $\langle \vec{w}, \vec{b} \rangle$ indicates the direction of the association (e.g., male or female, depending on the order of the word pairs). For higher dimensional bias subspaces, the sign of the projection cannot be interpreted in the same way. Also, as noted earlier, bias vectors are what are typically used to debias words in practice. As we show in the rest of this section, the interpretability of RIPA, its robustness to how the relation vector is defined, and its derivation from a method that provably debiases word embeddings are the key reasons why it is an ideal replacement for WEAT. Given that RIPA can be used for any embedding model that does matrix factorization, it is applicable to common embedding models such as SGNS and GloVe.

\subsection{Interpreting RIPA} If only a single word pair $(x,y)$ defines the association, then the relation vector $\vec{b} = (\vec{x} - \vec{y}) / \| \vec{x} - \vec{y} \|$, making RIPA highly interpretable. Given that RIPA is intended for embedding models that factorize a matrix $M$ containing a co-occurrence statistic (e.g., the shifted word-context PMI matrix for SGNS), if we assume that there is no reconstruction error, we can rewrite $\beta(\vec{w}; \vec{b})$ in terms of $M$. Where $x$ and $y$ have context vectors $\vec{x_c}$ and $\vec{y_c}$, $\lambda \in \mathbb{R}$ is such that $C = \lambda W$ (under the geometry-preserving conditions of Lemma 2 of \citet{ethayarajh2018towards}), $\alpha \in \mathbb{R}^-$ is a model-specific constant, and there is no reconstruction error:
\begin{equation}
    \begin{split}
        \beta_{\text{SGNS}}(\vec{w}; \vec{b}) &= \frac{(1/\lambda) \langle \vec{w}, \vec{x_c} - \vec{y_c} \rangle}{\| \vec{x} - \vec{y} \|} \\
        &= \frac{(1/\lambda)(\text{PMI}(x,w) - \text{PMI}(y,w))}{\sqrt{(1/\lambda)(-\text{csPMI}(x,y) + \alpha)}} \\
        &= \frac{1/\sqrt{\lambda}}{\sqrt{-\text{csPMI}(x,y) + \alpha}} \log \frac{p(w|x)} {p(w|y)}
    \end{split}
    \label{bias_directional}
\end{equation}
Here, $\text{csPMI}(x,y) \triangleq \text{PMI}(x,y) + \log p(x,y)$ and is equal to $-\lambda\| \vec{x} - \vec{y} \|^2_2 + \alpha$ under perfect reconstruction \citep{ethayarajh2018towards}. There are three notable features of this result:
\begin{enumerate}
    \item \citet{ethayarajh2018towards} proved the conjecture by \citet{pennington2014glove} that a word analogy holds over a set of words pairs $(x,y)$ iff for every word $w$, $\log [p(w|x)/ p(w|y)]$ is the same for every word pair $(x,y)$. The expression in (\ref{bias_directional}) is a multiple of this term.
    \item Assuming no reconstruction error, if a linear word analogy holds over a set of ordered word pairs $(x,y)$, then the co-occurrence shifted PMI (csPMI) should be the same for every word pair \citep{ethayarajh2018towards}. The more $x$ and $y$ are unrelated, the closer that $\text{csPMI}(x,y)$ is to $- \infty$ and $\beta(\vec{w};\vec{b})$ is to 0. This prevents RIPA from overestimating the extent of the association simply because $x$ and $y$ are far apart in embedding space.
    \item Because $\vec{b}$ is a unit vector, $\beta(\vec{w}; \vec{b})$ is bounded in $[ - \|\vec{w} \|, \| \vec{w} \| ]$. This means that one can calculate a word's association with respect to multiple relation vectors and then compare the resulting RIPA values.
\end{enumerate}
These points highlight just how robust RIPA is to the definition of $\vec{b}$. As long as a word analogy holds over the word pairs that define the association -- i.e., as long as the word pairs have roughly the same difference vector -- the choice of word pair does not affect $\log[p(w|x)/ p(w|y)]$ or $\text{csPMI}(x,y)$. Using (`king', `queen') instead of (`man', `woman') to define the gender relation vector, for example, would have a negligible impact. In contrast, as shown in section 4, the lack of robustness of WEAT to the choice of attribute sets is one reason it is so unreliable.

We can also interpret $\beta(\vec{w}; \vec{b})$ for other embedding models, not just SGNS. Where $X_{x,y}$ denotes the frequency of a word pair $(x,y)$ and $z_x, z_y$ denote the learned bias terms for GloVe:
\begin{equation}
    \begin{split}
        \beta_{\text{GloVe}} (\vec{w}; \vec{b}) &= C \left( \log \frac{p(x,w)}{p(y,w)} - z_x + z_y \right) \\
        \text{where}\ C &= \frac{1/\sqrt{\lambda}}{\sqrt{-\text{csPMI}(x,y) + \alpha}} \\
    \end{split}
\end{equation}
Because the terms $z_x, z_y$ are learned, $\beta(\vec{w}; \vec{b})$ is not as interpretable for GloVe. However, \citet{levy2015improving} have conjectured that, in practice, $z_x, z_y$ may be equivalent to the log counts of $x$ and $y$ respectively, in which case $\beta_{\text{GloVe}} = \beta_{\text{SGNS}}$.

\subsection{Statistical Significance} 

Unlike with WEAT, there is no notion of statistical significance attached to RIPA. There is a simple reason for this. Whether a word vector $\vec{w}$ is spuriously or non-spuriously associated with respect to a relation vector $(\vec{x} - \vec{y})/\|\vec{x} - \vec{y}\|$ depends on how frequently $(w,x)$ and $(w,y)$ co-occur in the training corpus; the more co-occurrences there are, the less likely the association is spurious. As shown in experiments by \citet{ethayarajh2018towards}, the reconstruction error for any word pair $(x,y)$ follows a zero-centered normal distribution where the variance is a decreasing function of $X_{x,y}$. Word embeddings alone are thus not enough to ascribe a statistical significance to the association. This also suggests that the notion of statistical significance in WEAT is disingenuous, as it ignores how the spuriousness of an association depends on co-occurrence frequency in the training corpus.

\begin{table*}[t]
    \footnotesize
    \begin{tabularx}{\textwidth}{cXccc}
        \toprule Word Type & Word & Genderedness in Corpus & Genderedness in Embedding Space & Change (abs.) \\ \midrule
   
        & mom & $-$0.163 & $-$0.648 & \ \ \ 0.485 \\
        & dad & \ \ \ 0.125 & \ \ \ 0.217 & \ \ \ 0.092 \\
        Gender-Appropriate & queen & $-$0.365 & $-$0.826 & \ \ \ 0.462 \\
         (n = 164) & king & \ \ \ 0.058 & \ \ \ 0.200 & \ \ \ 0.142 \\
        & \textbf{Avg (abs.)} & \ \ \ \textbf{0.231} & \ \ \ \textbf{0.522} & \ \ \ \textbf{0.291} \\ \midrule
        
        & nurse & $-$0.190 & $-$1.047 & \ \ \ 0.858 \\
        & doctor & $-$0.135 & $-$0.059 & $-$0.077 \\
        Gender-Biased & housekeeper & $-$0.132 & $-$0.927 & \ \ \ 0.795 \\
        (n = 68) & architect & $-$0.063 & \ \ \ 0.162 & \ \ \ 0.099 \\
        & \textbf{Avg (abs.)}  & \ \ \ \textbf{0.253} & \ \ \ \textbf{0.450} & \ \ \ \textbf{0.197} \\ \midrule
        
        & ballpark & \ \ \ 0.254 & \ \ \ 0.050 & $-$0.204 \\
        & calf & $-$0.039 & \ \ \ 0.027 & $-$0.012 \\
        Gender-Neutral & hormonal & $-$0.326 & $-$0.551 & \ \ \ 0.225 \\
        (n = 200) & speed & \ \ \ 0.036 & $-$0.005 & $-$0.031 \\
        & \textbf{Avg (abs.) } & \ \ \ \textbf{0.125} & \ \ \ \textbf{0.119} & \textbf{$-$0.006} \\ \bottomrule
        
    \end{tabularx}
    \caption{On average, SGNS makes gender-appropriate words (e.g., `queen') and gender-biased words (e.g., `nurse') \emph{more} gendered in the embedding space than they are in the training corpus. As seen in the last column (in bold), the average change in absolute genderedness is 0.291 and 0.197 respectively ($p < 0.001$ for both). For gender-neutral words, the average change is only $-$0.006 ($p = 0.84$): SGNS does not make them any more gendered.}
    \label{tab:breakdown}
\end{table*}

\section{Experiments}

With our experiments, we address two open questions. For one, how much of the gender association in an embedding space is due to the embedding model itself, how much is due to the training corpus, and how much is just noise? Secondly, how can we debias gender-biased words (e.g., `doctor', `nurse') but not gender-appropriate ones (e.g., `king', `queen') without \emph{a priori} knowledge of which words belong in which category?

\subsection{Setup}

For our experiments, we use SGNS embeddings trained on Wikipedia, since RIPA is highly interpretable for SGNS (see section 5.1). This means that for any given word in the vocabulary, we can compare its gender association in the training corpus to its gender association in the embedding space, which should be equal under perfect reconstruction. Words are grouped into three categories with respect to gender: \emph{biased}, \emph{appropriate}, and \emph{neutral}. We create lists of biased and appropriate words using the \citet{bolukbasi2016man} lists of gender-biased and gender-appropriate analogies. For example, \emph{doctor:nurse::man:woman} is biased, so we classify the first two words as biased. The last category, \emph{neutral}, contains uniformly randomly sampled words that appear at least 10K times in the corpus and that are not in either of the other categories, and which we therefore expect to be gender-agnostic.

\subsection{Breaking down Gender Association}

\begin{figure*}
    \centering
    \minipage{0.50\textwidth}
      \includegraphics[width=\linewidth]{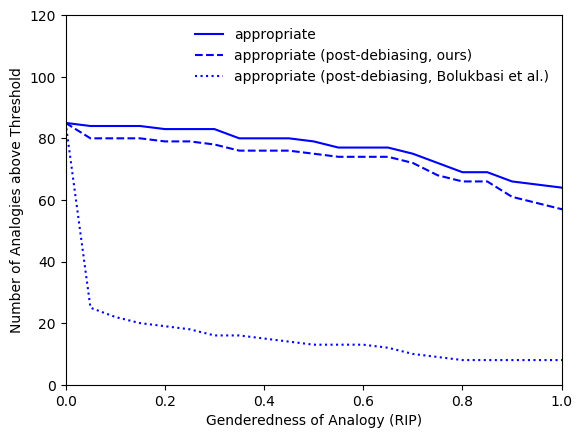}
    \endminipage\hfill
    \minipage{0.50\textwidth}%
      \includegraphics[width=\linewidth]{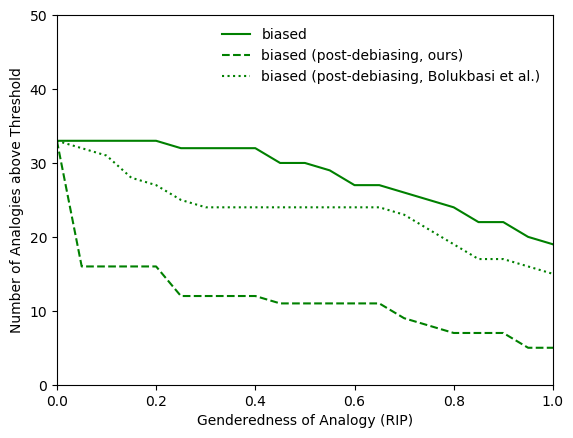}
    \endminipage
    \caption{Before debiasing words using subspace projection, one needs to identify which words are gender-appropriate -- to avoid debiasing them. The \citet{bolukbasi2016man} method of identifying these words is ineffective: it ends up precluding most gender-appropriate analogies (dotted line, left) while preserving most gender-biased analogies (dotted line, right). Our unsupervised method (dashed line) does much better in both respects. }
    \label{fig:analogies}
\end{figure*}

For any given word, under the conditions above, the gender association in the training corpus is what the gender association in the embedding space would be if there were no reconstruction error. By comparing these two quantities, we can infer the change induced by the embedding model. Let $g(w;x,y)$ denote the RIPA of a word $w$ with respect to the gender relation vector defined by word pair $(x,y)$, let $\hat{g}(w;x,y)$ denote what $g(w;x,y)$ would be under perfect reconstruction for an SGNS embedding model, and let $\Delta_{g}$ denote the change in absolute gender association from corpus to embedding space. Where $S$ is a set of gender-defining word pairs\footnote{ \footnotesize The set of gender-defining pairs we used is \{(`woman', `man'), (`girl', `boy'), (`she', `he'), (`mother', `father'), (`daughter', `son'), (`gal', `guy'), (`female', `male'), (`her', `his'), (`herself', `himself'), (`mary', `john')\}.} from \citet{bolukbasi2016man} and $\lambda, \alpha$ are the model-specific constants defined in section 5.1,
\begin{equation}
\begin{split}
    g(w;x,y) &= \frac{\langle \vec{w}, \vec{x} - \vec{y} \rangle}{\| \vec{x} - \vec{y}\|} \\
    \hat{g}(w;x,y) &=  \frac{1/\sqrt{\lambda}}{\sqrt{-\text{csPMI}(x,y) + \alpha}} \log \frac{p(w|x)}{p(w|y)} \\
    \Delta_{g}(w;S) &= \left| \sum_{(x,y) \in S} \frac{g(w;x,y)}{|S|} \right| - \left| \sum_{(x,y) \in S} \frac{\hat{g}(w;x,y)}{|S|} \right| \\
\end{split}
\label{change_in_bias}
\end{equation}
We take the absolute value of each term because the embedding model may make a word more gendered, but in the direction opposite of what is implied in the corpus. $\lambda \gets 1$ because we expect $\lambda \approx 1$ in practice \citep{ethayarajh2018towards, mimno2017strange}. Similarly, $\alpha \gets -1$ because it minimizes the difference between $\| \vec{x} - \vec{y} \|$ and its information theoretic interpretation over the gender-defining word pairs in $S$, though this is an estimate and may differ from the true value of $\alpha$. In Table \ref{tab:breakdown}, we list the gender association in the training corpus ($g(w)$), the gender association in embedding space ($\hat{g}(w)$), and the absolute change $(\Delta_g(w))$ for each group of words.

On average, the SGNS embedding model does not make gender-neutral words any more gendered than they are in the training corpus. Given that much of the vocabulary falls into this category, this means that the embedding model does not systematically change the genderedness of most words. However, because of reconstruction error, individual words may be more or less gendered in the embedding space, simply due to chance. In contrast, for words that are either gender-biased or gender-appropriate, on average, the embedding model actually amplifies the gender association in the corpus. For example, for the word `king', which is gender-specific by definition, the association is $0.058$ in the corpus and $0.200$ in the embedding space -- it becomes more male-associated. For the word `nurse', which is gender-biased, the association is $-0.190$ in the corpus and $-1.047$ in the embedding space -- it becomes more female-associated. On average, the amplification is much greater for gender-appropriate words than it is for gender-biased ones, although the latter are more gendered in the corpus itself. In both cases, the change in absolute genderedness is statistically significant ($p < 0.001$).

This amplification effect is unsurprising and can be explained by second-order similarity. Two words can be nearby in a word embedding space if they co-occur frequently in the training corpus (first-order similarity) or if there exists a large set of context words with which they both frequently co-occur (second-order similarity). The latter explains why words like `Toronto' and `Melbourne' are close to each other in embedding space; both are cities that appear in similar contexts. In an environment with some reconstruction error, such as low-dimensional embedding spaces, second-order similarity permits words to be closer in embedding space than would be the case if only first-order similarity had an effect. As a result, $\lambda \langle \vec{\textit{king}}, \vec{\textit{man}} \rangle > (\text{PMI}(\text{`king'}, \text{`man'}) - \log k)$ for SGNS, for example. What is often treated as a useful property of word embeddings can have, with respect to gender bias, a pernicious effect.

\subsection{Debiasing without Supervision}

To use the subspace projection method \citep{bolukbasi2016man}, one must have prior knowledge of which words are gender-appropriate, so that they are not debiased. Debiasing all vectors can preclude gender-appropriate analogies such as \emph{king:queen::man:woman} from holding in the embedding space. To create an exhaustive list of gender-appropriate words, \citet{bolukbasi2016man} started with a small, human-labelled set of words and then trained an SVM to predict more gender-appropriate terms in the vocabulary. This bootstrapped list of gender-appropriate words was then left out during debiasing.

The way in which \citet{bolukbasi2016man} evaluated their method is unorthodox: they tested the ability of their debiased embedding space to generate new analogies. However, this does not capture whether gender-appropriate analogies are successfully preserved and gender-biased analogies successfully precluded. In Figure \ref{fig:analogies}, we show how the number of appropriate and biased analogies changes after debiasing. The x-axis captures how strongly gendered the analogy is, using the absolute RIPA value $|\beta(\vec{w};\vec{b})|$ but replacing $\vec{w}$ with the difference vector defined by the first word pair (e.g., $\vec{\textit{king}} - \vec{\textit{queen}}$). The y-axis captures the number of analogies that meet that threshold. 

As seen in Figure \ref{fig:analogies}, \citeauthor{bolukbasi2016man}'s bootstrapped list of gender-appropriate words yields the opposite of what is intended: it ends up precluding most gender-appropriate analogies and preserving most gender-biased ones. This is not the fault of the debiasing method; rather, it is the result of failing to correctly identify which words in the vocabulary are gender-appropriate. For example, the bootstrapped list\footnote{Available at https://github.com/tolga-b/debiaswe} includes `wolf_cub' and `Au_Lait' as gender-appropriate terms, even though they are not. Conversely, it fails to include common gender-appropriate words such as `godfather'. This problem highlights how finding the right words to debias is as important as the debiasing itself. 

We propose an unsupervised method for finding gender-appropriate words. We first create a gender-defining relation vector $\vec{b}^*$ by taking the first principal component of gender-defining difference vectors such as $\vec{\textit{man}} - \vec{\textit{woman}}$. Using difference vectors from biased analogies, such as $\vec{\textit{doctor}} - \vec{\textit{midwife}}$, we then create a bias-defining relation vector $\vec{b}'$ the same way. We then debias a word $w$ using the subspace projection method iff it satisfies $|\beta(\vec{w}; \vec{b}^*)| < |\beta(\vec{w}; \vec{b}')|$. As seen in Figure \ref{fig:analogies}, this simple condition is sufficient to preserve almost all gender-appropriate analogies while precluding most gender-biased ones. 

In our debiased embedding space, 94.9\% of gender-appropriate analogies with a strength of at least 0.5 are preserved in the embedding space while only 36.7\% of gender-biased analogies are. In contrast, the \citet{bolukbasi2016man} approach preserves only 16.5\% of appropriate analogies with a strength of at least 0.5 while preserving 80.0\% of biased ones. Recall that we use the same debiasing method as \citet{bolukbasi2016man}; the difference in performance can only be ascribed to how we choose the gender-appropriate words. Combining our heuristic with other methods may yield even better results, which we leave as future work.


\section{Conclusion}
In this paper, we answered several open questions about undesirable word associations in embedding spaces. We found that for any embedding model that implicitly does matrix factorization (e.g., SGNS, GloVe), debiasing with the subspace projection method is, under certain conditions, equivalent to training on a corpus that is unbiased with respect to the words defining the bias subspace. We proved that WEAT, the most common test of word embedding association, has theoretical flaws that cause it to systematically overestimate bias. For example, by contriving the attribute sets for WEAT, virtually any word can be classified as gender-biased relative to another. We then derived a new measure of association in word embeddings called the relational inner product association (RIPA). Using RIPA, we found that SGNS does not, on average, make most words any more gendered in the embedding space than they are in the training corpus. However, for words that are gender-biased or gender-specific by definition, SGNS amplifies the genderedness in the corpus.

\section*{Acknowledgments}

We thank the anonymous reviewers for their insightful comments. We thank the Natural Sciences and Engineering Research Council of Canada (NSERC) for their financial support.
This version of the paper clarifies some assumptions in Lemma 2 of \citet{ethayarajh2018towards}; the rest is unchanged from the published version.

\bibliography{naaclhlt2019}

\begin{thebibliography}{14}
\expandafter\ifx\csname natexlab\endcsname\relax\def\natexlab#1{#1}\fi

\bibitem[{Arora et~al.(2016)Arora, Li, Liang, Ma, and
  Risteski}]{arora2016latent}
Sanjeev Arora, Yuanzhi Li, Yingyu Liang, Tengyu Ma, and Andrej Risteski. 2016.
\newblock A latent variable model approach to {PMI}-based word embeddings.
\newblock \emph{Transactions of the Association for Computational Linguistics},
  4:385--399.

\bibitem[{Bengio et~al.(2003)Bengio, Ducharme, Vincent, and
  Jauvin}]{bengio2003neural}
Yoshua Bengio, R{\'e}jean Ducharme, Pascal Vincent, and Christian Jauvin. 2003.
\newblock A neural probabilistic language model.
\newblock \emph{Journal of Machine Learning Research}, 3(Feb):1137--1155.

\bibitem[{Bolukbasi et~al.(2016)Bolukbasi, Chang, Zou, Saligrama, and
  Kalai}]{bolukbasi2016man}
Tolga Bolukbasi, Kai-Wei Chang, James~Y Zou, Venkatesh Saligrama, and Adam~T
  Kalai. 2016.
\newblock Man is to computer programmer as woman is to homemaker? debiasing
  word embeddings.
\newblock In \emph{Advances in Neural Information Processing Systems}, pages
  4349--4357.

\bibitem[{Caliskan et~al.(2017)Caliskan, Bryson, and
  Narayanan}]{caliskan2017semantics}
Aylin Caliskan, Joanna~J Bryson, and Arvind Narayanan. 2017.
\newblock Semantics derived automatically from language corpora contain
  human-like biases.
\newblock \emph{Science}, 356(6334):183--186.

\bibitem[{Collobert and Weston(2008)}]{collobert2008unified}
Ronan Collobert and Jason Weston. 2008.
\newblock A unified architecture for natural language processing: Deep neural
  networks with multitask learning.
\newblock In \emph{Proceedings of the 25th International Conference on Machine
  Learning}, pages 160--167. ACM.

\bibitem[{Ethayarajh et~al.(2018)Ethayarajh, Duvenaud, and
  Hirst}]{ethayarajh2018towards}
Kawin Ethayarajh, David Duvenaud, and Graeme Hirst. 2018.
\newblock Towards understanding linear word analogies.
\newblock \emph{arXiv preprint arXiv:1810.04882}.

\bibitem[{Gonen and Goldberg(2019)}]{gonen2019lipstick}
Hila Gonen and Yoav Goldberg. 2019.
\newblock Lipstick on a pig: Debiasing methods cover up systematic gender
  biases in word embeddings but do not remove them.
\newblock \emph{arXiv preprint arXiv:1903.03862}.

\bibitem[{Landauer and Dumais(1997)}]{landauer1997solution}
Thomas~K Landauer and Susan~T Dumais. 1997.
\newblock A solution to {Plato's} problem: The latent semantic analysis theory
  of acquisition, induction, and representation of knowledge.
\newblock \emph{Psychological review}, 104(2):211.

\bibitem[{Levy and Goldberg(2014)}]{levy2014neural}
Omer Levy and Yoav Goldberg. 2014.
\newblock Neural word embedding as implicit matrix factorization.
\newblock In \emph{Advances in Neural Information Processing Systems}, pages
  2177--2185.

\bibitem[{Levy et~al.(2015)Levy, Goldberg, and Dagan}]{levy2015improving}
Omer Levy, Yoav Goldberg, and Ido Dagan. 2015.
\newblock Improving distributional similarity with lessons learned from word
  embeddings.
\newblock \emph{Transactions of the Association for Computational Linguistics},
  3:211--225.

\bibitem[{Mikolov et~al.(2013)Mikolov, Sutskever, Chen, Corrado, and
  Dean}]{mikolov2013distributed}
Tomas Mikolov, Ilya Sutskever, Kai Chen, Greg~S Corrado, and Jeff Dean. 2013.
\newblock Distributed representations of words and phrases and their
  compositionality.
\newblock In \emph{Advances in Neural Information Processing Systems}, pages
  3111--3119.

\bibitem[{Mimno and Thompson(2017)}]{mimno2017strange}
David Mimno and Laure Thompson. 2017.
\newblock The strange geometry of skip-gram with negative sampling.
\newblock In \emph{Proceedings of the 2017 Conference on Empirical Methods in
  Natural Language Processing}, pages 2873--2878.

\bibitem[{Pennington et~al.(2014)Pennington, Socher, and
  Manning}]{pennington2014glove}
Jeffrey Pennington, Richard Socher, and Christopher Manning. 2014.
\newblock {GloVe}: Global vectors for word representation.
\newblock In \emph{Proceedings of the 2014 Conference on Empirical Methods in
  Natural Language Processing (EMNLP)}, pages 1532--1543.

\bibitem[{Zhao et~al.(2018)Zhao, Zhou, Li, Wang, and Chang}]{zhao2018learning}
Jieyu Zhao, Yichao Zhou, Zeyu Li, Wei Wang, and Kai-Wei Chang. 2018.
\newblock Learning gender-neutral word embeddings.
\newblock In \emph{Proceedings of the 2018 Conference on Empirical Methods in
  Natural Language Processing}, pages 4847--4853.

\end{thebibliography}
\bibliographystyle{acl_natbib}
\clearpage

\appendix
\end{document}